\RequirePackage{fix-cm}
\documentclass[smallextended]{svjour3}

\smartqed  

\usepackage{graphicx}
\usepackage{amsmath}
\usepackage{amssymb}
\usepackage{algorithmic}
\usepackage{algorithm}
\usepackage{subfigure}
\usepackage{array}
\usepackage{tabularx}
\usepackage{multirow}
\usepackage{color}

\newcommand{\bfx}{{\textbf{x}}}
\newcommand{\bfv}{{\textbf{v}}}
\newcommand{\bfu}{{\textbf{u}}}

\newcommand{\bfxi}{{\boldsymbol{\xi}}}

\begin{document}

\title{Sparse Coding with Earth Mover's Distance for Multi-Instance Histogram Representation}

\author{ Mohua Zhang \and Jianhua Peng \and Xuejie Liu \and Jim Jing-Yan Wang}

\institute{
M. Zhang and J. Peng\at
National Digital Switching System Engineering \& Technology Research Center, Zhengzhou 450002, HeNan, China
\\
\email{mohua.zhang@outlook.com}
\and
M. Zhang\at
College of Computer and Information Engineering, Henan University of Economics and Law, Zhengzhou 450002, HeNan, China
\and
X. Liu and J. Wang\at
University at Buffalo, The State University of New York, Buffalo, NY 14260, USA\\
\email{xuejie.liu@hotmail.com}
}

\date{Received: date / Accepted: date}

\maketitle

\begin{abstract}
Sparse coding (Sc) has been studied very well as a powerful data representation method. It attempts to represent the feature vector of a data sample by reconstructing it as the sparse linear combination of some basic elements, and a $L_2$ norm distance function is usually used as the loss function for the reconstruction error. In this paper, we investigate using Sc as the representation method within multi-instance learning framework, where a sample is given as a bag of  instances, and further represented as a histogram of the quantized instances. We argue that for the data type of histogram, using $L_2$ norm distance is not suitable, and propose to use the earth mover's distance (EMD) instead of $L_2$ norm distance as a measure of the reconstruction error. By minimizing the EMD between the  histogram of a sample and the its reconstruction from some basic histograms, a novel sparse coding method is developed, which is refereed as SC-EMD. We evaluate its performances as a histogram representation method in tow multi-instance learning problems --- abnormal image detection in wireless capsule endoscopy videos, and protein binding site retrieval. The encouraging results demonstrate the advantages of the new method over the traditional method using $L_2$ norm distance.
\end{abstract}

\section{Introduction}

Sparse Coding (SC) has been recently proposed and studied well as an effective data representation method in machine learning community \cite{Lee2007,Yang20091794,Mairal201019,Yang201545,Bai2015,Representing2015}. Given a set of basic elements and a data sample, SC tries to represent the sample by reconstructing it as a linear combination of these
basic elements. The linear combination coefficient vector could be used as the new representation of the sample.
To this end, the basic elements and the coefficient vector (also called coding vector) are learned by minimizing the reconstruction error. At the same time, we also hope that the coding vector could be as sparse as possible.
To measure the reconstruction error, a squared $L_2$ norm distance is usually applied to
compare the original feature vector and its sparse linear combination
as a loss function. At the same time, a $L_1$ norm  regularization term is also imposed to
the coding vector to seek its sparsity.
The advantage of using the $L_2$ norm  distance
to as the loss function
and using $L_1$ norm regularization for sparsity purpose
lies on that it is easy to optimize and interpret.
The feature-sign search method had been proposed
to solve the SC problem by Lee et al. in \cite{Lee2007}.
Some different SC versions had also be proposed since then, by adding
different bias terms to the original SC loss function based on $L_2$ norm  distance \cite{LapSC2010,GraphSC2011,Gao2013}.

In the multi-instance learning framework, each sample is given
as a bag of multiple instances, instead of one single instance in the traditional
machine learning problem \cite{Zhou20091249,Zhou2005135,Zhou20071609,Du201551}.
For example, in image classification and retrieval problems,
an image could be split into many small image patches, and each patch is an instance.
In this case, we usually first learn a set of instance prototypes
by clustering the instances of the training samples, and
then represent a sample by quantizing its instances into the
instance prototypes, and obtain a quantization histogram \cite{Tsai2009100,Lu2005956,Kotani2002II105}.
The normalized histogram is used as the feature vector of the sample
for further classification or retrieval task.
When we try to apply the SC to represent the histogram data samples under the
multi-instance learning framework,
directly using $L_2$ norm distance may not be suitable anymore.
Other distance functions which is especially suitable for histogram data is desired.
In fact, many distance functions have been studied for histogram data comparison,
such as Kullback --- Leibler divergence \cite{Seghouane200797,Rached2004917,Hershey2007IV317}, $\chi^2$ distance \cite{Huong20091310,Emran200219,Ye2006393}, Earth Mover's Distance (EMD) \cite{Levina2001251,Ling2007840,Rubner200099}, etc.
Among these distance functions, the EMD metric has been
known to quantify the errors in histogram comparison better than other distance metrics.

In this paper, we propose the first SC method with
EMD metric for the representation of histogram data.
Instead of using $L_2$ norm distance, we model the
sparse coding problem by using the EMD to constructed the loss function.
The newly proposed method, SC-EMD is especially suitable for the representation of histogram data
under the multi-instance learning framework.

This rest parts of this paper continue as follows:  the
formal objective function of SC-EMD, the linear programming-based optimization,
and an iterative learning algorithm, are presented in
section \ref{sec:SCEMD};
experiments with three actual multi-instance learning
tasks are presented in section \ref{sec:exp}; and
conclusions are given in section \ref{sec:conclusion}.

\section{Sparse Coding with Earth Mover's Distance}
\label{sec:SCEMD}

In this section, we will introduce the novel
sparse coding method using EMD as the distance metric, instead of
the traditional squared $L_2$ norm distance, for the
representation of histogram data.

\subsection{Objective function}

Assuming that we have a training set of $N$ data samples.
We represent each data sample as a bag of multiple instances under the framework of multi-instance learning.
To extract the feature vector from the $n$-th sample,
we quantize the instances of the $n$-th sample into
a set of instance prototypes, and use the quantization histogram as the
feature vector.
We assume that the number of instance prototypes is $D$, thus the feature vector of the $n$-th sample is
a normalized $D$-dimensional histogram, denoted as  $\bfx_n=[x_{n1}, \cdots x_{nD}]^\top \in  \mathbb{R}_+^{D}$, where $x_{ni}$ is the $i$-th bin of the histogram.
Note that $\bfx_n$ is normalized as $\sum_{i=1}^D x_{ni}=1$, so that it is a distribution.
The set of the histograms of all the training samples is denoted as $\mathcal{X}=\{\bfx_1,\cdots,\bfx_N\}$,
where $\bfx_n$ is the histogram of the $n$-th sample.

Under the framework of sparse coding, we try to represent each histogram in $\mathcal{X}$
as a sparse linear combination of a set of basic histograms.
We denote the set of basic histograms as $\mathcal{U}=\{\bfu_1,\cdots,\bfu_M\}$,
where $\bfu_m=[u_{m1}, \cdots u_{mD}]^\top \in \mathbb{R}_+^D$
is the $m$-th basic histogram, and $M$ is the number of basic histograms.
Similar to $\bfx_n$, $\bfu_m$ is also normalized as  $\sum_{i=1}^D h_{mi}=1$.
The basic histograms are further organized as a  basic matrix
$U=[\bfu_1,\cdots,\bfu_M] \in  \mathbb{R}_+^{D\times M}$.
With the basic histograms, we try to reconstruct each $\bfx_n$ as the weighted linear combination of these basic histograms, as

\begin{equation}
\begin{aligned}
\bfx_n \approx \sum_m \bfu_m v_{nm}
= U \bfv_n,
\end{aligned}
\end{equation}
where $\bfv_n = [v_{n1},\cdots, v_{nM}]^\top \in \mathbb{R}^M$
is the reconstruction coefficient vector, which is also called coding vector of $\bfx_n$.
$v_{nm}$ is the coefficient of the $m$-th basic histogram for the reconstruction
of the histogram of the $n$-th sample.
Similarly, the sparse coding vectors for all the samples in $\mathcal{X}$ could also be
organized as a coding matrix as $V=[\bfv_1,\cdots,\bfv_N] \in  \mathbb{R}^{M\times N}$,
with the $n$-th column as the coding vector of the $n$-th sample.
Given the histogram $\bfx_n$ of $n$-th sample, the target of sparse coding problem is  to learn
a basic histogram matrix $U$, and a sparse coding vector $\bfv_n$,
so that the original $\bfx_n$ and its reconstruction $U \bfv_n$ should be as
close to each other as possible, and the reconstruction error can be minimized.
At the same time, we also expect the coding vector $\bfv_n$ to be as sparse as possible.
To this end, we discuss the following two issues to build the objective function for the
learning of $U$ and $V$.

\begin{description}
\item[Reconstruction Error]
To measure the reconstruction error between $\bfx_n$ and $U \bfv_n$, traditional sparse coding
methods have used the squared $L_2$ norm distance, as

\begin{equation}
\begin{aligned}
L_2(\bfx_n, U \bfv_n)=
\left \|  \bfx_n - U \bfv_n \right \|^2_2,
\end{aligned}
\end{equation}
for the learning of $U$ and $V$. The objective function is built by applying the squared $L_2$ norm distance to all samples in the training set.
However, as we discussed in the introduction section,
$L_2$ norm distance is unsuitable for the histogram data.
In this work, we try to apply the EMD as a distance measure between $\bfx_n$ and $U \bfv_n$,
which has been a popular metric for histogram data.
To define the EMD  between two histograms $\bfx_n$ and $U \bfv_n$,
we treat each bin of $\bfx_n$ as a supply, while each bin of $U \bfv_n$ as a demand.
We also denote $d_{ij}$ as the ground distance from the $i$-th supply to $j$-th demand.
The EMD between $\bfx_n$ and $U \bfv_n$ is defined as
the minimum amount of work needed to fill all the demands with all the supplies,

\begin{equation}
\label{equ:EMD}
\begin{aligned}
EMD(\bfx_n, U \bfv_n)=
&
\underset{F^n}{\min}
\sum_{i,j}f^n_{ij} d_{ij}\\
&
s.t. f^n_{ij}\geq 0,  \sum_j f^n_{ij} \leq x_{ni},  \sum_i f^n_{ij} \leq \sum_m u_{mj} v_{nm},
\end{aligned}
\end{equation}
where variable $f^n_{ij}$ denotes the amount transported from the $i$-th supply to the $j$-th
demand for the $n$-th sample, and $F^n=[f^n_{ij}]^{D\times D}_+$ is the matrix of the transported amounts.
The constrain $f^n_{ij}\geq 0$ prevents the negative transportation.
The constrain $\sum_j f^n_{ij} \leq x_{ni}$ means that the mess moved out from the $i$-th supply should not
be larger than $x_{ni}$, while $\sum_i f^n_{ij} \leq \sum_m u_{mj} v_{nm}$ means that the mess moved into the $j$-th demand should not be larger than $\sum_m u_{mj} v_{nm}$.
The problem in (\ref{equ:EMD}) could be solved as a Linear Programming (LP) problem \cite{Liu2014397,BenTal2000411,Candes20054203}.

\item[Sparsity Regularization]
To encourage the sparsity of each coding vector $\bfv_n$, traditional sparse coding approaches
have been imposing a $L_1$ norm based sparsity penalty to $\bfv_n$ as

\begin{equation}
\begin{aligned}
\min_{\bfv_n}
\left\{
L_1(\bfv_n)=
||\bfv_n||_1=
\sum_{m=1}^{M}
|v_{nm}|
\right\}
\end{aligned}
\end{equation}
Using the $L_1$ norm sparsity penalty, we can impose most of the the
elements of $\bfv_n$ to zeros, and only a few of them will be kept for the
reconstruction of $\bfx_n$.

Direct optimization of this regularization term is difficulty, because it is non-convex and non-smooth. We follow the works of Fan et al. \cite{fan2014finding,fan2010enhanced,fan2011margin,fan2015improved,fan2014tightening} which are proposed to improve the lower bounds for Bayesian network structure learning, and propose to optimize the upper bound of the $L_1$ norm regularization. Fan et al. \cite{fan2014tightening} proposed a method to tighten the upper and lower bounds of the learning problem of the Bayesian network structure. In the work of  Fan et al. \cite{fan2014tightening}, more informed variable groupings are used to create the pattern databases for the tightening of the lower bounds, and an anytime learning algorithm is used for the tightening of the upper bounds. Moreover, Fan et al. \cite{fan2015improved} proposed a new partition method to use the information extracted from the potential optimal parent sets to improve the lower bound for Bayesian network structure learning. Inspired by the works of Fan et al. \cite{fan2014tightening,fan2015improved}, to solve the problem together with the LP problem of EMD, instead of minimizing the $L_1$ norm of the code vector $\bfv_n$ directly, we introduce a slack vector the upper bound of its $L_1$ norm, and minimize its $L_1$ norm. We first introduce a nonnegative slack vector $\bfxi_n\in \mathbb{R}^{M}_+$ for each code vector as the upper bound of the absolute vector of the code vector, $\bfxi_n\geq |\bfv_n| \geq 0$, where $|\bfv_n|= [ |\bfv_{n1}|,\cdots,|\bfv_{nM}| ]\in \mathbb{R}^{M}_+$, and then minimize the $L_1$ norm of the slack vector to seek the sparsity of $\bfv_n$ indirectly.
Because $\bfxi_n$ is a nonnegative vector, its $L_1$ norm could be computed simply as the summation of its elements as $L_1(\bfxi_n)=\sum_{m=1}^M \xi_{nm}$. To seek the sparsity of $\bfv_n$, we have the following optimization problem,

\begin{equation}
\label{equ:slack}
\begin{aligned}
\min_{\bfxi_n,\bfv_n}
&
\left \{
L_1(\bfxi_n)=\sum_{m=1}^M \xi_{nm}
\right \}\\
s.t.
&
-\bfxi_n\leq \bfv_n \leq \bfxi_n,
\bfxi_n\geq 0.
\end{aligned}
\end{equation}
We also organize the upper bound vectors for the spare codes as a upper bound matrix
$\Xi=[\bfxi_1,\cdots,\bfxi_N]\in \mathbb{R}^{M\times N}_+$,
where the $n$-th column $\bfxi_n$ is the upper bound vector of the $n$-th sparse code vector.
By using the slack vectors, we make the sparsity regularization term a
smooth function, which could be integrated to the optimization problem
of EMD naturally, and could solved as  LP problem easily.

\end{description}

By applying both the EMD based reconstruction error term in (\ref{equ:EMD}) and the
sparsity regularization term in (\ref{equ:slack}) to each training sample in
$\mathcal{X}$, and summing them up, we have the following objective function
for the EMD based sparse coding problem,

\begin{equation}
\label{equ:SCEMD1}
\begin{aligned}
\underset{U,V,\Xi}{\min}
&\sum_{n} \left( EMD(\bfx_n, U \bfv_n) + \gamma L_1 (\bfxi_n)\right)\\
s.t.
& u_{mj}\geq 0, \sum_j u_{mj} =1,\\
&-\bfxi_n\leq \bfv_n \leq \bfxi_n,~
\bfxi_n\geq 0,
\end{aligned}
\end{equation}
where $\gamma$ is a trade-off parameter, and the constrains $u_{mj}\geq 0$ and $\sum_j u_{mj} =1$
are introduced to the basic histograms to guarantee that the learned basic histograms are normalized distributions.
Please note that the $EMD(\bfx_n, U \bfv_n)$ itself is also obtained by solving
a minimizing problem with regarding to $F^1,\cdots,F^M$. We
substitute (\ref{equ:EMD}) and (\ref{equ:slack}) to (\ref{equ:SCEMD1}), so that the optimization
problem in (\ref{equ:SCEMD1}) is extended
into the parameter-enlarged optimization
with additional parameters of transported amount matrices as,

\begin{equation}
\label{equ:SCEMD2}
\begin{aligned}
\underset{U,V,\Xi, F^n|_{n=1}^N}{\min}
&
\sum_{n}
\left (
\sum_{i,j}f^n_{ij} d_{ij}
+ \gamma \sum_{m}\xi_{nm}
\right )
\\
s.t.
& u_{mj}\geq 0, \sum_j u_{mj} =1,\\
& f^n_{ij}\geq 0,\sum_j f^n_{ij} \leq x_{ni}, \sum_i f^n_{ij} \leq \sum_m u_{mj} v_{nm},\\
&-\xi_{nm} \leq v_{nm} \leq \xi_{nm},~
\xi_{nm}\geq 0.
\end{aligned}
\end{equation}
This problem is a parameter-enlarged LP problem.

\subsection{Optimization}

Directly optimizing the object of (\ref{equ:SCEMD2}) is difficult and time-consuming.
Similar to the original $L_2$ norm-based sparse coding method, we adopt an alternate optimization strategy
for the learning of $U$ and $(V,\Xi)$ in an iterative algorithm.
In each iteration, one of $U$ and $(V,\Xi)$  will be optimized while the other one is
fixed, and then their roles will be switched.
The iteration will be repeated until a maximum iteration number is reached.

\subsubsection{Optimizing $(V,\Xi)$ while fixing $U$}

By fixing $U$, we could optimize the coding vectors in $V$ together with
the other additional variables. Similar to the traditional sparse coding methods, we update each sparse coding vector individually. When the coding vector $\bfv_n$ of the $n$-th sample and its slack vector $\bfxi_n$ are being optimized, the other ones $\bfv_{n'} (n'\neq n)$ with their corresponding  additional variables ($F^{n'}$ and $\bfxi_{n'}$) are fixed.
Thus, the optimization problem in (\ref{equ:SCEMD2}) will be turned to

\begin{equation}
\label{equ:SCEMDvn}
\begin{aligned}
\underset{\bfv_n, \bfxi_n, F^n}{\min}
&
\left (
\sum_{i,j}f^n_{ij} d_{ij}
+ \gamma   \sum_m \xi_{nm}
\right )
\\
s.t.
& f^n_{ij}\geq 0,\sum_{j} f^n_{ij} \leq x_{ni}, \sum_i f^n_{ij} \leq \sum_m u_{mj} v_{nm},\\
& -\xi_{nm} \leq v_{nm} \leq \xi_{nm}, \xi_{nm}\geq 0,
\end{aligned}
\end{equation}
which could be solved as a LP problem. The LP problem is solve by using a active set algorithm. Please notice that LP solves a problem for a given vector of unknown variables. Here we substitute the vector of variables in $F^n$ (the original variables of the EMD problem) with a longer vector, which contains the entries in $F^n$, $\bfv_n$, and $\bfxi_{n}$.  We did not reformulate the EMD objective, but we changed its constraints to take care of the additional variables related to the the problem solved. In this way, the new LP problem is different from that of the original EMD, and the result contains entries of $F^n$, $\bfv_n$ and $\bfxi_{n}$.

\subsubsection{Optimizing $U$ while fixing $(V,\Xi)$}

By fixing $V$ and $\Xi$, the optimization problem in (\ref{equ:SCEMD2}) can be turned to

\begin{equation}
\label{equ:SCEMDU}
\begin{aligned}
\underset{U,F^n|_{n=1}^N}{\min}
&
\sum_{n}
\sum_{i,j}f^n_{ij} d_{ij}
\\
s.t.
& \sum_j u_{mj} =1, u_{mj}\geq 0,\\
& f^n_{ij}\geq 0,\sum_{j} f^n_{ij} \leq x_{ni}, \sum_i f^n_{ij} \leq \sum_m u_{mj} v_{nm}.
\end{aligned}
\end{equation}
which could also be solved as a LP problem using active set algorithm.

An important limitation of both the optimization problems in (\ref{equ:SCEMDvn}) and (\ref{equ:SCEMDU})
is the large number of additional variables for the LP problem.
For each sample $\bfx_n$, a $D\times D$
transported amount matrix $F^n$ is solved in both (\ref{equ:SCEMDvn}) and (\ref{equ:SCEMDU}),
thus there are totally $N \times D \times D$ transportation amount variables in the
LP problem for the $N$ training samples. When the dimension of the histogram $D$, or the training sample $N$ is large,
there would be a large number of variables, which could cause serious computation problem.
To overcome this shortage, we reduce the number of variables in $F^n$ by allowing moving the earth
from the $i$-th supply only to its $K$ nearest demands instead of all the $D$ demands.
The $K$ nearest demands of $i$-th supply is found by using the ground distances.
In this way, we reduce the transported mass variables for each supply of each sample from $D$ to $K$.
Usually $K\ll D$, thus the total transported amount is reduced significantly from $N \times D \times D$
to $N \times D \times K$.

\subsection{Algorithm}

We summarize  the iterative basic histograms and coding vectors learning algorithm in Algorithm \ref{alg:SCEMD}. In each iteration, the sparse coding vector for each sample is first learned sequentially,
and the basic histograms are then updated based on the learned sparse coding vectors.
The iterations will be repeated $T$ times.
When a novel sample comes with its histogram in the test procedure,
we simply solve (\ref{equ:SCEMDvn}) to obtain its sparse coding vector.

\begin{algorithm}[h!]
\caption{SC-EMD Algorithm.}
\begin{algorithmic}
\label{alg:SCEMD}
\STATE \textbf{Input}: Histograms of training samples $\mathcal{X}=\{\bfx_i,\cdots, \bfx_N\}$;
\STATE \textbf{Input}: Number of basic histograms $M$;
\STATE \textbf{Input}: Maximum number of iterations $T$.

\STATE Initialize the basic histogram matrix  $U^0=\bfu_i^0,\cdots, \bfu_M^0$;

\FOR{$t=1,\cdots,T$}
\FOR{$n=1,\cdots,N$}
\STATE Update the sparse coding vector $\bfv_n^t$ for the $n$-th sample
by solving (\ref{equ:SCEMDvn}) while fixing $U^{t-1}$;
\ENDFOR
\STATE Update the basic histogram matrix  $U^t$ by solving (\ref{equ:SCEMDU})
while fixing $V^{t}$;
\ENDFOR
\STATE \textbf{Output}: The basic histogram matrix $U^T$ and the sparse coding matrix $V^{T}$.
\end{algorithmic}
\end{algorithm}

\subsection{Relation to nonnegative matrix factorization with earth mover's distance}

Sandler and Lindenbaum \cite{sandler2011nonnegative} proposed nonnegative matrix factorization with earth mover's distance (NMFEMD), which factorize a nonnegative matrix by minimizing the EMD between the original data matrix and the product of two factorization matrices. Our work SC-EMD  has close relation to it. We discuss the relations of the two methods as following:
\begin{itemize}
\item Both SC-EMD and NMFEMD use earth mover's distance to measure the reconstruction error of the data, which is a suitable distance measure for multi-instance quantization histogram.
\item NMFEMD dose impose the sparsity of the factorization matrices, while SC-EMD impose the reconstruction coefficients to be sparse. The sparsity of the reconstruction coefficients is measure by a $L_1$ norm. Thus the objective function of SC-EMD is different compared to NMFEMD, because the objective of NMFEMD is only a EMD term, while the object of SC-EMD is composed of a EMD term and a $L_1$ term. The optimization of SC-EMD is more difficult than NMFEMD due to this additional term.
\item NMFEMD imposes the reconstruction matrices to be nonnegative, while SC-EMD doesn't have such constraints. However, these constraints do not change the optimization of the objective. The NMFEMD is optimized as a linear constrained LP problem. Adding the nonnegative constraint only adds some more linear constraints to the problem. But SC-EMD adds a $L_1$ norm regularization term to the objective, and changes the optimization problem.
\end{itemize}

\section{Experiments}
\label{sec:exp}

In this section, we evaluated the proposed method on three multi-instance learning
problems, where each feature vector is a histogram for each sample.

\subsection{Experiment I: Abnormal Image
Detection in Wireless Capsule Endoscopy Videos}

Wireless Capsule Endoscopy (WCE) has been used to
detect the mucosal abnormalities
in the gastrointestinal tract, including  blood, ulcer, polyp, etc \cite{Iddan2000417,Ell2002685,Mylonaki20031122,Hwang2011320}.
However, usually only a few frames from a large number of WCE videos
contain abnormalities, thus a medical clinician spends long time to find the abnormal frames from a WCE
video. In this situation, it is very necessary to develop a system to automatically
discriminate abnormal frames from the normal ones.
In this experiment, we evaluated the proposed method as image representation
method for the task of abnormal image detection in WCE videos.

\subsubsection{Dataset and Setup}

We constructed the data set for the experiment by collecting 170 images of WCE videos belonging to three abnormal classes and one normal class.
The data set contains 50 normal images, 40 polyp images, 40 ulcer images, and 40 blood images.
Given an image of WCE video, the task of abnormal image detection is to classify it to one of the four classes. To this end, each image was split into many $8\times 8$ small patches, and each patch was treated as
an instance, thus the image was represented as a bag of instances under the framework of multi-instance learning. We extract color and texture features from each patch and concatenate them
as visual features of each instance.
Then the instances were quantized into a pool of instance prototypes
and the quantization histogram was normalized and used as the feature vector of the image.
The histograms were further represented using the proposed SC-EMD algorithm
as the sparse coding vectors, and the coding vectors were used to train a
Support Vector Machine (SVM) \cite{Cherkassky2004113,Schuldt200432,Keerthi2001637,Tsang2005} to classify the images into one of the four image types.

To conduct the experiment, we employed the 10-fold cross-validation protocol \cite{Gandek19981171,Craven1978377,Bagby199423}.
The entire data set was split into 10 non-overlapping folds randomly.
In each fold, there were 5 normal images, 4 polyp images, 4 ulcer images, and 4 blood images respectively.
Each fold was used as the test set in turn, and the remaining 9 folds are combined and used as the training set. After the images in the training set were represented as histograms under the multi-instance learning framework, we performed the SC-EMD algorithm to them and obtain the basic histograms and the sparse coding vectors for the training images.
Then we train a SVM classifier from these sparse coding vectors for each class.
To handle the multi-class problem, we used the one-against-all protocol to train classifiers \cite{Kumar201114238,Liu2005849,Polat20091587}.
A SVM classifier was trained for each class, using the images of this class as positive samples, while all other images as negative samples.
Based on the basic histograms learned from training histograms, we represented the test images
and obtain the sparse coding vectors, and finally input them into the learned SVM classifiers to have the final classification results. Please note that the parameters were turned using only the training set while excluding the test set.

The classification results are measured by the recall-precision curve \cite{Goadrich2006231,Gordon1989145,Huang20051665}, Receiver Operating Characteristic (ROC) curve \cite{Cook2007928,DeLong1988837,Hanley198229} and the Area Under the ROC Curve (AUC) value \cite{Pencina2008157,DeLong1988837,Hanley198229} for each class.

\subsubsection{Results}

\begin{figure}[htb!]
\centering
\subfigure[Blood]{
\includegraphics[width=0.48\textwidth]{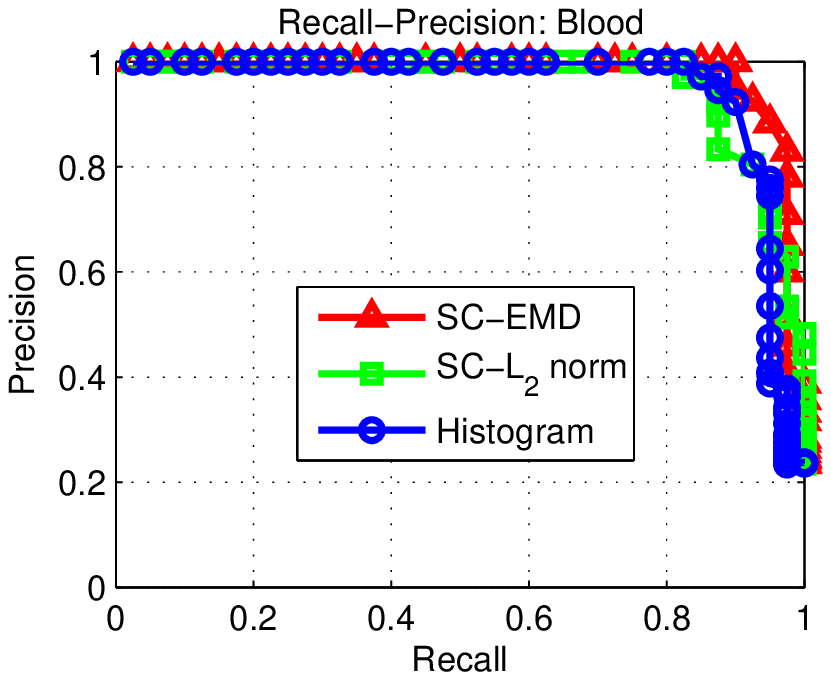}}
\subfigure[Polyp]{
\includegraphics[width=0.48\textwidth]{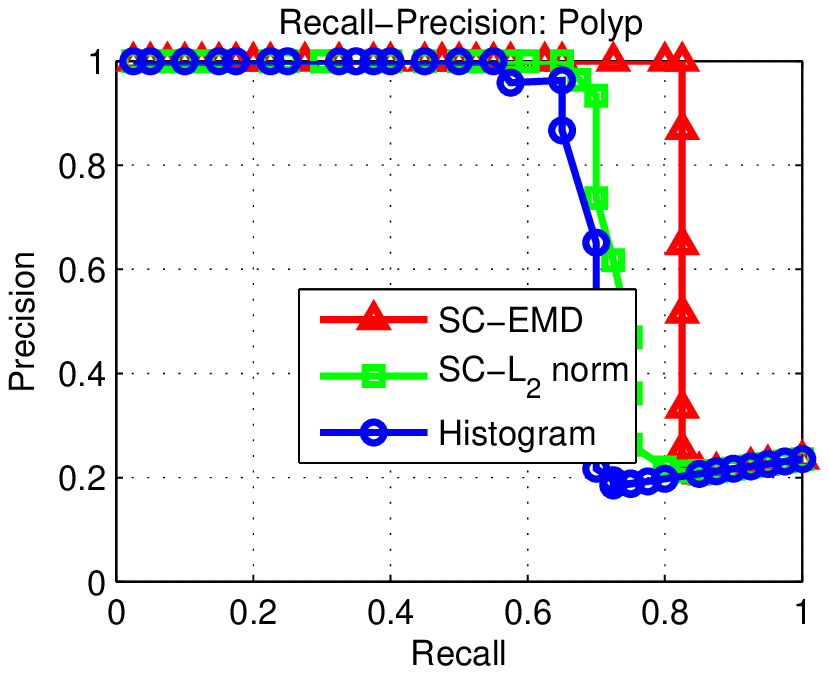}}\\
\subfigure[Ulcer]{
\includegraphics[width=0.48\textwidth]{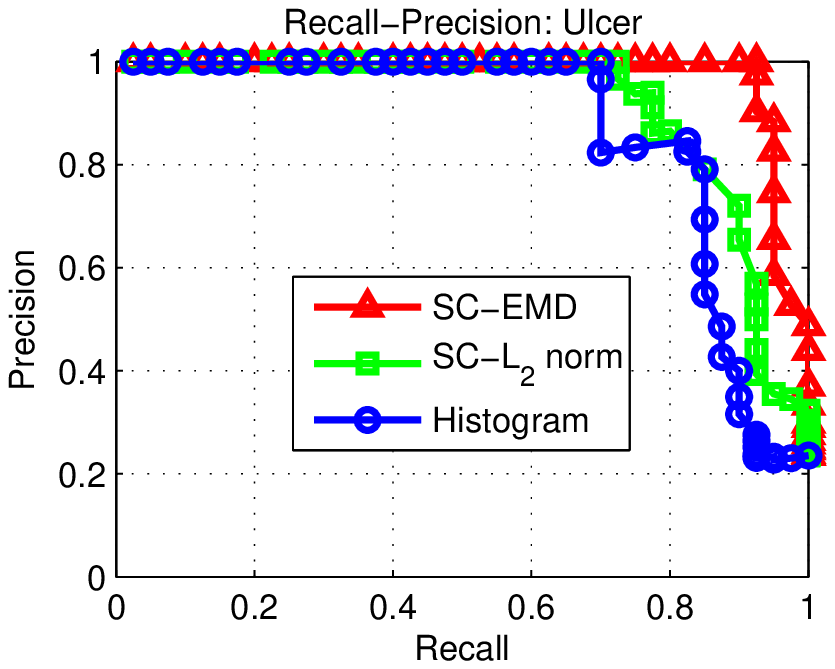}}
\subfigure[Normal]{
\includegraphics[width=0.48\textwidth]{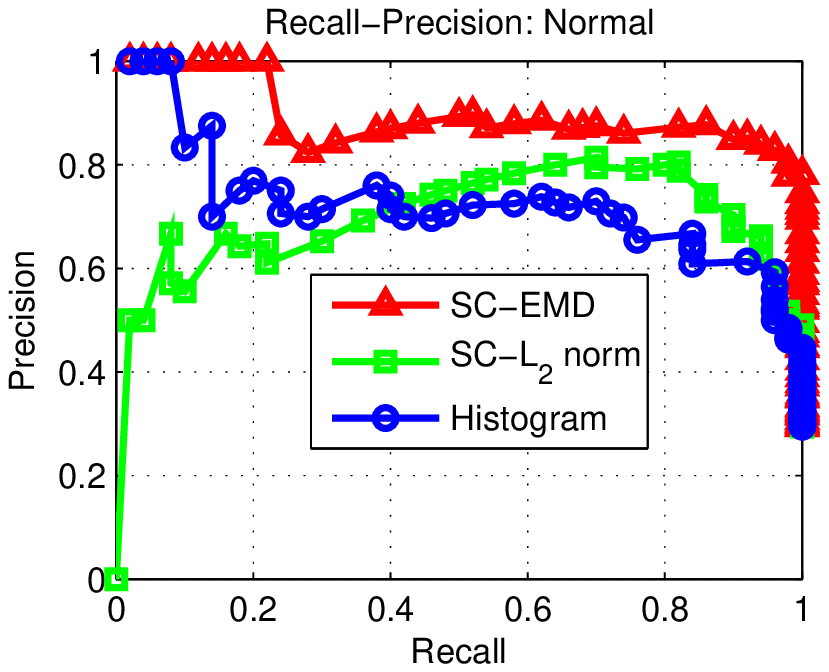}}
\caption{The recall-precession curves of different classes using different histogram representation methods on the
WCE images database.}
\label{fig:FigWCERecPre}
\end{figure}

In the experiments, we compared our SC-EMD algorithm
as a data representation method against the traditional  sparse coding
method using the $L_2$ norm distance (denoted as SC-$L_2$ norm),
and also against the original histogram as representation (denoted as Histogram).
The recall-precision curves for four different classes are given in
Fig. \ref{fig:FigWCERecPre} (a) - (d).
In these figures, it is clearly
shown that with the proposed SC-EMD, the classification performances for all
four classes are improved significantly, even more so for the last three classes.
The performance improvement is particularly dramatic for the Polyp and Normal classes.
SC-$L_2$ norm could improve the original histogram features somehow, however, due to the
reason that it employs the $L_2$ norm distance as loss function, which is not suitable for
the histogram data, the improvement is limited.
In particular, Fig. \ref{fig:FigWCERecPre} (a) shows that an
increase in classification performance is obtained
by SC-$L_2$ norm against both original histogram and SC-EMD. The
results validate the importance of performing sparse coding with appropriate loss function to the histogram data.

\begin{figure}[htb!]
\centering
\subfigure[Blood]{
\includegraphics[width=0.48\textwidth]{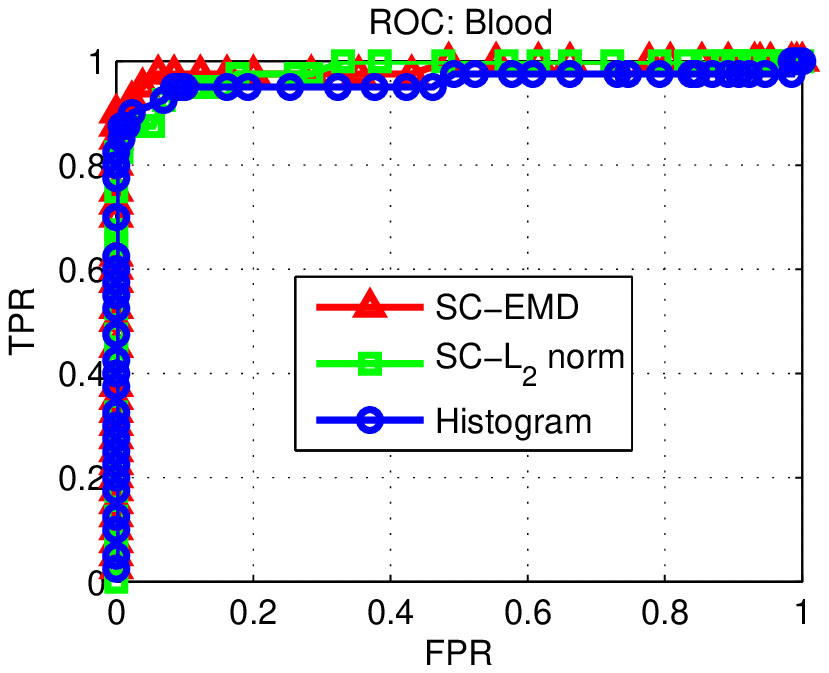}}
\subfigure[Polyp]{
\includegraphics[width=0.48\textwidth]{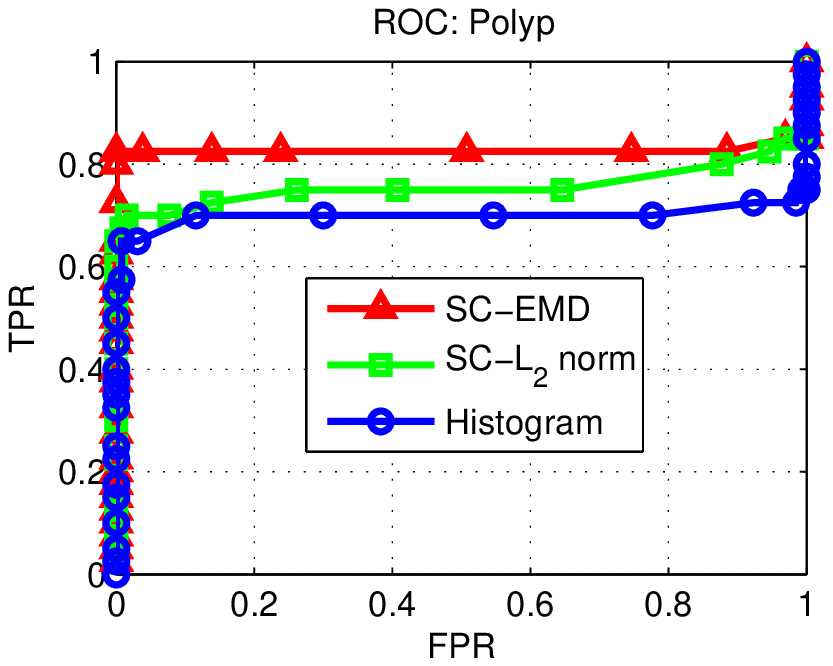}}\\
\subfigure[Ulcer]{
\includegraphics[width=0.48\textwidth]{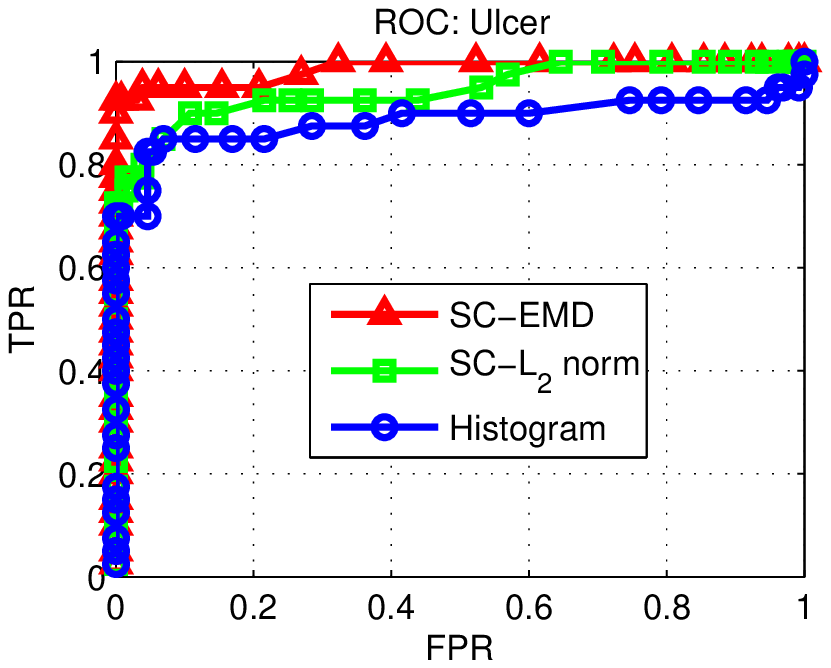}}
\subfigure[Normal]{
\includegraphics[width=0.48\textwidth]{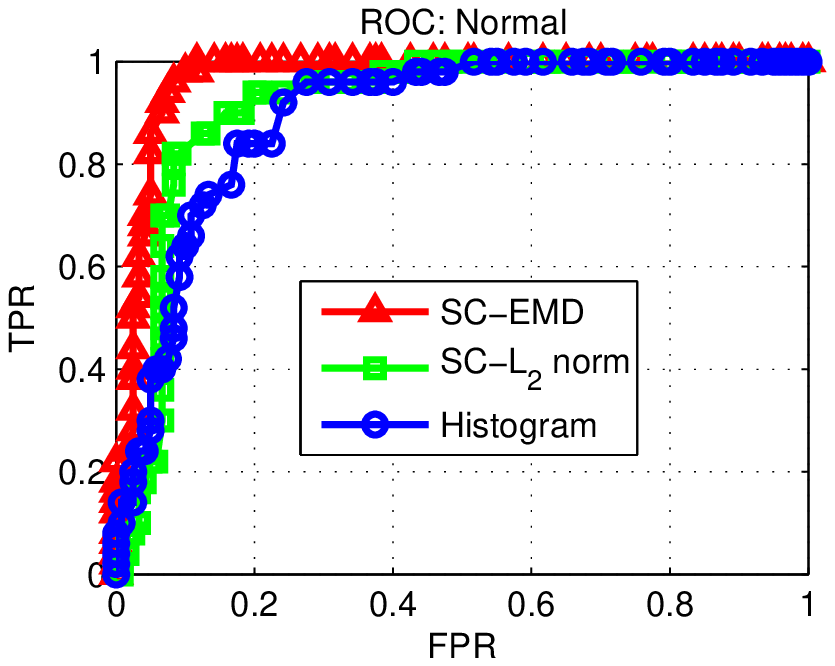}}
\caption{The ROC curves of different classes using different histogram representation methods on the
WCE images database.}
\label{fig:FigWCEROC}
\end{figure}

\begin{figure}[htb!]
\centering
\includegraphics[width=0.7\textwidth]{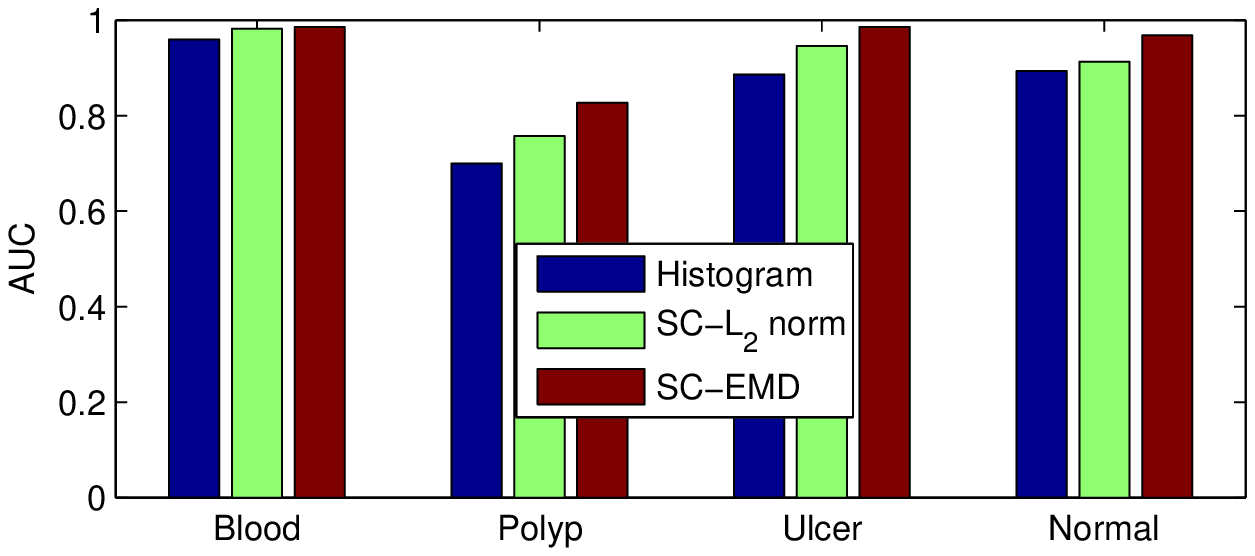}
\caption{The AUC valuse of different classes using different histogram representation methods on the
WCE images database.}
\label{fig:FigWCEAUC}
\end{figure}

The ROC curves of different classes are shown in
Fig. \ref{fig:FigWCEROC}. Moreover, the AUC values are given in Fig. \ref{fig:FigWCEAUC}.
{As shown in Fig. \ref{fig:FigWCEROC} and Fig. \ref{fig:FigWCEAUC}, our
SC-EMD algorithm clearly outperforms the original histogram feature and
SC-$L_2$ norm-based method in all four classes again.
The advantage is particularly significant on the
more challenging normal class. This result
highlights the importance of using the EMD measure for histograms rather
than $L_2$-norm distance.}

\subsection{Experiment II: Protein binding site retrieval}

Searching geometrically similar protein binding sites is
significantly important to understand the functions of protein and also to
drug discovery \cite{Bind2012,Bradford20051487,Neuvirth2004181,Laurie20051908}.
Pang et al. \cite{Bind2012} presented the protein binding sites
as a histogram using the multi-instance learning framework for the protein binding site retrieval problem.
In this experiment, we evaluated the proposed algorithm for the representation of
histograms of protein binding sites.

\subsubsection{Dataset and Setup}

\begin{figure}[htb!]
\centering
\includegraphics[width=\textwidth]{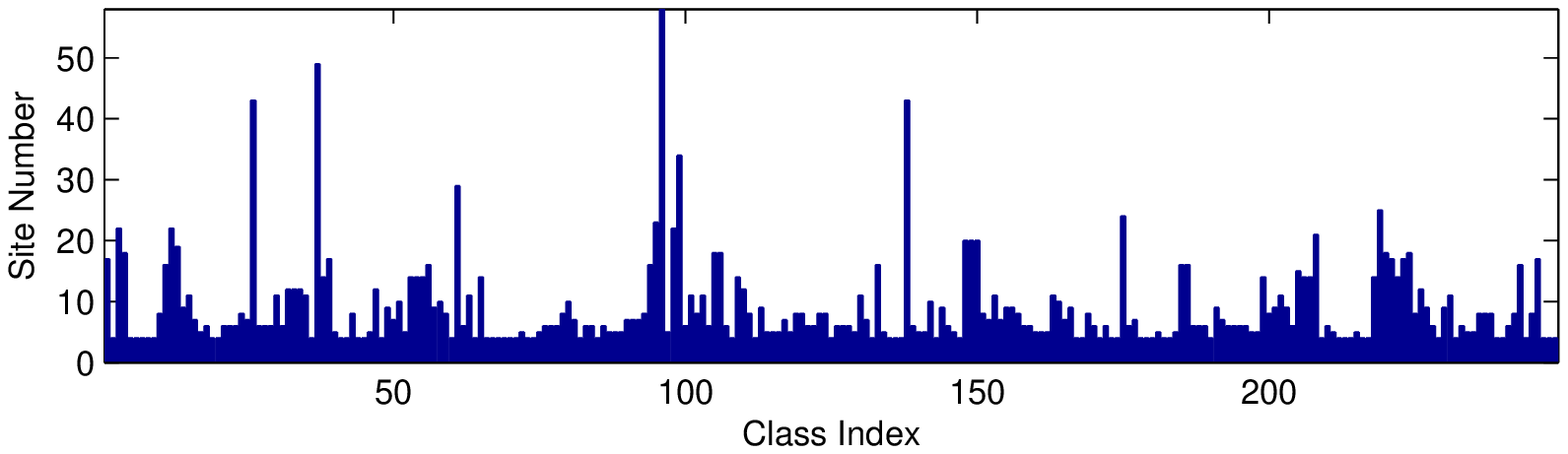}
\caption{The numbers of sites in different classes of the protein binding site dataset.}
\label{fig:FigBindData}
\end{figure}

In this experiment, we used a protein binding site data set reported by  Pang et al. \cite{Bind2012}.
In this non-redundant data set, there are totally 2,819 protein binding sites, belonging to 501 different classes.
The number of sites in each class varies from 2 to 58.
To conduct the 4-fold cross-validation, we had selected 2,226
binding sites randomly to construct our data set.
The selected data set contained sites of 249 classes, and
the number of sites for each class was from 4 to 58, so that we could guarantee that
when the 4-fold cross-validation was performed, in each fold there were at least one
site from every class.
The numbers of sites for all the selected classes are shown in Fig. \ref{fig:FigBindData}.

Given a query
binding site and a protein binding site database,
the protein binding site retrieval problem is to
rank the database sites according to their similarity
to the query in a descending order, so that the database sites belonging to the same class as the query can be ranked at the top positions of the returned list. To this end, we first represented each binding site as a bag of feature points selected from the binding site surface, and for each point the geometric features were extracted \cite{Bind2012}.
In the multi-instance learning framework, a binding site was refereed as a bag, and each feature point was refereed as a instance. Then all the feature points were quantized into a set of prototype points, and a histogram was generated as
the bag-level feature of the binding site \cite{Bind2012}.
Using the proposed SC-EMD algorithm, the histograms are represented as sparse codes for the final ranking.
The ranking performances were evaluated by the recall-precision and the ROC curves.
AUC values of the ROC curve were also reported as a single performance measure of the ranking results.

\subsubsection{Results}

\begin{figure}[htb!]
\centering
\subfigure[Recall-Precision]{
\includegraphics[width=0.48\textwidth]{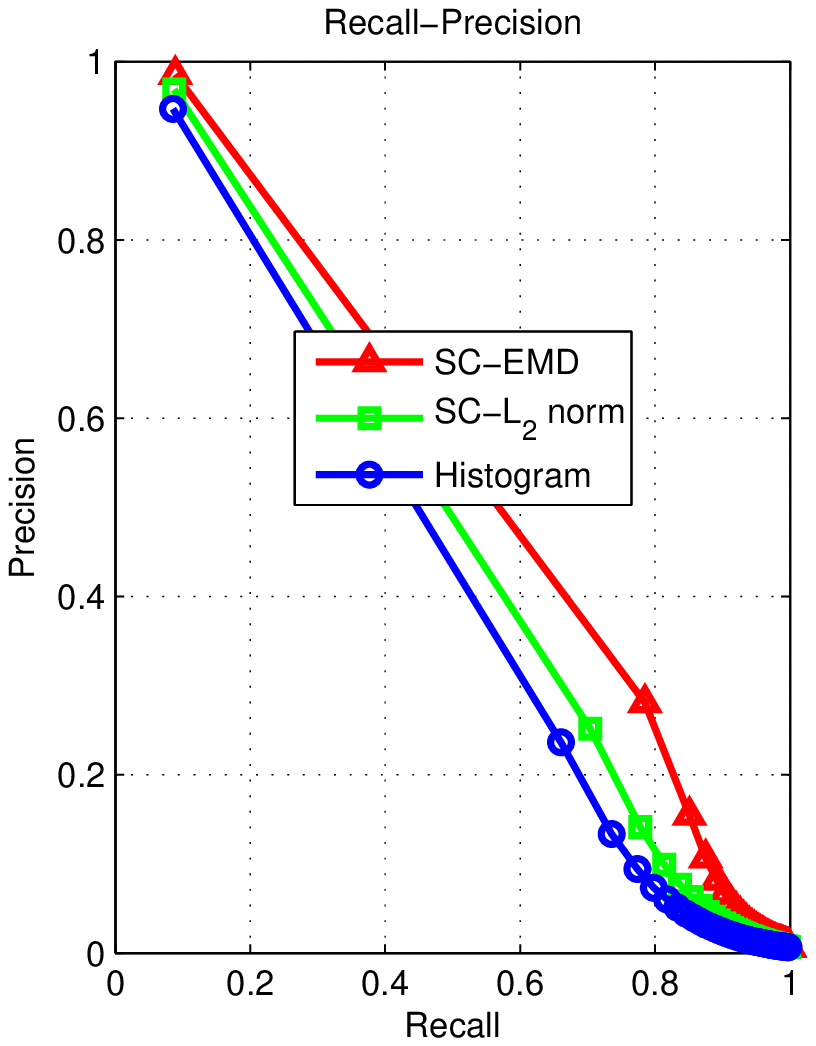}}
\subfigure[ROC]{
\includegraphics[width=0.48\textwidth]{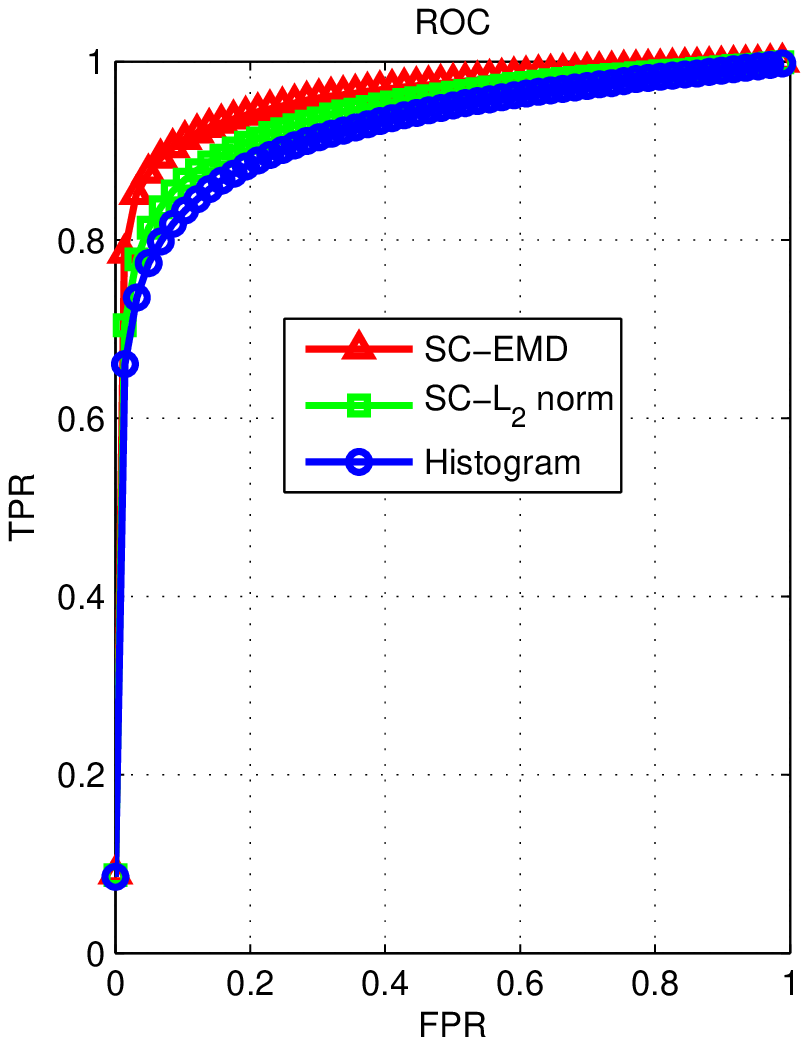}}
\caption{The recall-precision and ROC curves of different histogram representation methods on the
protein binding site database.}
\label{fig:FigBind}
\end{figure}

\begin{figure}[htb!]
\centering
\includegraphics[width=0.8\textwidth]{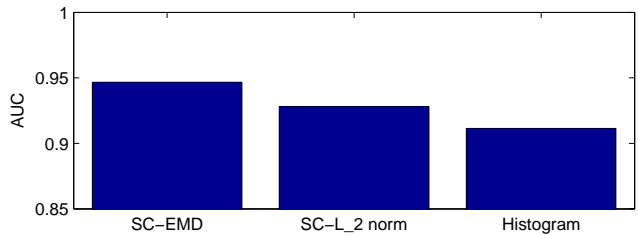}
\caption{The AUC values of ROC curves of different histogram representation methods on the
protein binding site database.}
\label{fig:FigBindAUC}
\end{figure}

The recall-precision and ROC curves of different histogram representation methods
are given in Fig. \ref{fig:FigBind}.
From the results in Fig. \ref{fig:FigBind}, we can find that the SC-EMD method performs the best
in terms of both recall-precision and ROC curves. It proves that the EMD based method could discover the best  distance measure for histogram comparison and coding. The AUC values of the ROC curves are also given in Fig. \ref{fig:FigBindAUC}. These protein binding site retrieval system with SC-EMD representation method achieves
an AUC value of 0.9466, compared to an AUC value of 0.9282 using SC-$L_2$ norm and 0.9114 using the original histogram.

\subsection{Experiment III: Object Recognition}

In this section, we compared the proposed method to some other feature extraction and classification methods
on an publicly accessed image database.

\subsubsection{Dataset and Protocol}

In this experiment, we used the COREL-2000 image database which is popular in
the computer vision community for the problem of object recognition \cite{Tang2007384,chen2004image}.
In this data set, there were 2,000 images of 20 objects. For each object, the number of images was
100. The target of image recognition was to assign a given image to a class of object correctly.

To conduct the experiment, we also used the 10-fold cross-validation. Each images was regarded as a sample, and we extracted the Regions Of Interest (ROI) as the instances \cite{Christopoulos2000247,Heikkil2009425}. We extracted multiple ROIs for each image, thus each image was represented as a bag of multiple instances.
Moreover, the instances of the training images were clustered to generate a set of instance prototypes
using a clustering algorithm \cite{Frey2007972,Xu2005645,Xie1991841,Jain1999316},
and then the instances of each image were quantized into it to present each image as a histogram.
The histograms were firstly normalized and then the proposed SC-EMD algorithm was applied to
represent the histograms to sparse codes. The histogram of the training samples were used
to learn the basic histograms and their sparse codes first, and then a
SVM classifier was learned in the sparse code space for each class.
To classify a test sample, we also represented its histogram to a sparse code vector using the basic histograms learned from the training set, and then classified the sparse code vector using the SVM classifier learned from the training set.
To evaluate the classification performances of the  proposed algorithm, we used the
classification accuracy as the performance measure \cite{Lim2000203,Baldi2000412,Foody2002185},
which is computed as,

\begin{equation}
\label{equ:accuracy}
\begin{aligned}
accuracy=\frac{Number~ of~ correctly~ classified~test~ images}{Number~ of~ test~ images}.
\end{aligned}
\end{equation}

\subsubsection{Results}

\begin{figure}
\centering
\subfigure[Comparison to feature selection methods]{
\includegraphics[height=0.3\textheight]{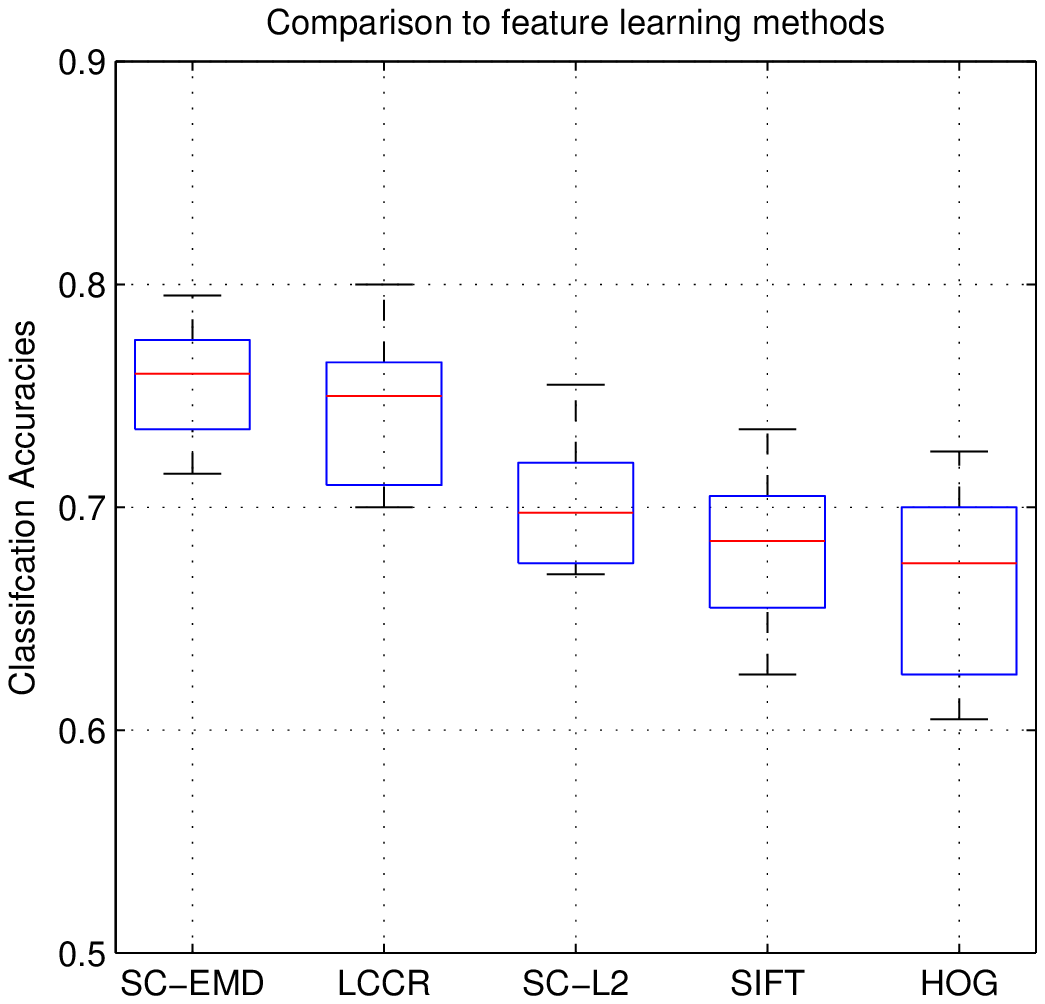}}
\subfigure[Comparison to classifiers]{
\includegraphics[height=0.3\textheight]{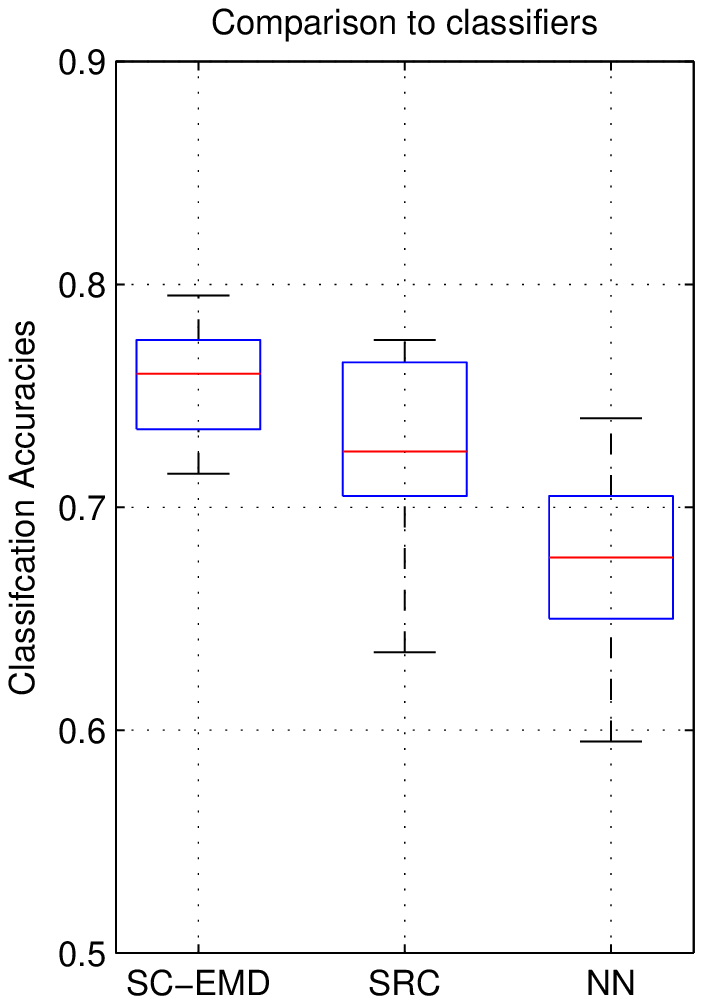}}\\
\caption{Experiment results on the COREL-2000 image data set.}
\label{fig:COREL}
\end{figure}

In this experiment, we compared the proposed histogram representation algorithm against several visual
feature extraction methods, including the Learning Locality-Constrained Collaborative Representation (LCCR) method proposed by Peng et al. \cite{Peng2013Learning}, SC-$L_2$,
Histogram of Oriented Gradients (HOG) \cite{Suard2006206,Zhu20061491,Dalal2005886},
and Scale-Invariant Feature Transform (SIFT) \cite{Li2007332,Shen20091714,Cheung20092012}.
The experiment results are given in Fig. \ref{fig:COREL} (a).
As we can see from this figure, the proposed method outperforms all the other methods significantly besides
LCCR. SIFT and HOG both represent the images as histograms, however, they ignore the structure of the data set by representing each images individually, thus the performances are inferior to others.
SC-$L_2$ explorers the training set by learning a set of basic histograms to represent all the image histograms, and it achieves some minor improvements. But it uses the $L_2$ norm distance to compare the histograms, which is not suitable.
It is very interesting to notice that LCCR, which improves the robustness and discrimination of data representation by introducing the local consistency, has archived similar performances to SC-EMD and outperformed other methods too.
Although it also uses the $L_2$ norm as loss function which is unsuitable for histograms, it
considers the local consistency of the data samples, and seek the smoothness of the code instead of the
sparsity. These are the main reasons for the good performances of LCCR. It also encourages us to develop
novel methods to combine EMD and locality-constrained collaborative representation, which may even improve the performance more significantly.
Moreover, we also compare our method to two popular classification methods, including
Sparse Representation-based Classification (SRC) \cite{Wright2009210},
Nearest Neighbor classification (NN) \cite{Zhang20062126,Weinberger2009207,Denoeux1995804}.
The results is given in Fig. \ref{fig:COREL} (b).
As we can see from the figure, the proposed algorithm based on EMD outperforms both the two classification using $L_2$ norm distance as distance metric for histograms.
This is another evidence that EMD is essential for histogram data analysis and classification.

\section{Conclusion and Future Works}
\label{sec:conclusion}

A new type of sparse coding method, sparse coding with EMD metric, is proposed in this paper
for the representation of histogram data.
The objective function is composed of an EMD term between the original histogram and the
reconstruction result from a pool of basic histograms,
and a $L_1$ term for the regularization of the coding vector.
The optimization problem is solved as  a LP problem in an iterative algorithm.
Algorithms based on the proposed SC-EMD outperformed previous $L_2$-norm based sparse coding algorithms
in three challenging multi-instance learning tasks.

\section*{Acknowledgements}

The work was support partly by the foundation for innovative research groups of the national natural science foundation of China (Grant No. 61521003), partly by science and technique foundation of  HeNan province, China (Grant No. 152102210087), partly by foundation of educational committee of HeNan province, China(Grant No. 14A520040).


\end{document}